\documentclass[letterpaper, 10pt, conference]{ieeeconf}      
\usepackage[utf8]{inputenc}
\IEEEoverridecommandlockouts                              

\usepackage[export]{adjustbox}
\usepackage{graphicx}
\usepackage{mathtools}
\usepackage{xfrac}
\usepackage[english]{babel}
\usepackage[T1]{fontenc}
\usepackage{amsmath,amssymb,bbm} 
\usepackage{array}
\usepackage{mathrsfs}
\usepackage{cite}
\usepackage{multirow}
\usepackage{amsmath}
\usepackage{subfig}
\usepackage{multirow}
\usepackage{color}
\usepackage{url}
\usepackage[table]{xcolor}
\newcommand{\squeezeup}{\vspace{-2.5mm}}

\title{\LARGE \bf Understanding Natural Language Instructions for Fetching Daily Objects Using GAN-Based Multimodal Target-Source Classification
}

\author{Aly Magassouba, Komei Sugiura, Anh Trinh Quoc and Hisashi Kawai
 \thanks{National Institute of Information and Communications Technology, 3-5 Hikaridai, Seika, Soraku, Kyoto 619-0289, Japan  {\tt\small name.surname@nict.go.jp}} 
 } 
 
\newcommand*\Update{\color{black}}
\newcommand*\Done{\color{black}}
\def\LW{\dimexpr.25\linewidth-1.5em}
\def\LWb{\dimexpr.05\linewidth-0.5em}

\begin{document}

\maketitle
\graphicspath{{figures/}} 
\begin{abstract}
In this paper, we address multimodal language understanding with unconstrained fetching instruction for domestic service robots. A typical fetching instruction such as ``Bring me the yellow toy from the white shelf'' requires to infer the user intention, {\it i.e.}, what object (target) to fetch and from where (source). To solve the task, we propose a Multimodal Target-source Classifier Model (MTCM), which predicts the region-wise likelihood of target and source candidates in the scene. Unlike other methods, MTCM can handle region-wise classification based on linguistic and visual features. We evaluated our approach that outperformed the state-of-the-art method on a standard data set. We also extended MTCM with Generative Adversarial Nets (MTCM-GAN), and enabled simultaneous data augmentation and classification. 
\end{abstract}

\section{Introduction}
The increasing demand for support services for older and disabled people has spurred the development of assistive robots \cite{brose2010role} as  an alternative and credible solution to the shortage of caring labor. 
Increasing efforts are being made to standardize domestic service robots (DSRs) that provide various support functions \cite{iocchi2015robocup}. 

Meanwhile, most DSRs still have a limited ability  to interact through language. Specifically, most DSRs do not allow users to instruct them with various expressions relating to object retrieval tasks. Communicative DSRs  will provide non-expert users with a user-friendly way to instruct DSRs.

Against this background, our work addresses multimodal language understanding for fetching instructions (MLU-FI). This task consists of predicting the target object as instructed by the user from an unconstrained sentence, such as ``{\it Please give me the empty bottle from the small white shelf}.'' Such a task raises several challenges related to the ambiguity of the instructions; {\it i.e.}, the relevant information may be truncated, hidden, or expressed in several ways. The many-to-many nature of mapping between the language and real world makes it difficult to accurately predict user intention. Consequently, an MLU-FI problem requires the fusion and exploitation of heterogeneous data before extracting an appropriate interpretation.

Dialogue systems \cite{gemignani2015language} can be used for disambiguation, as is  usually the case during DSRs competitions like RoboCup@Home\cite{iocchi2015robocup}. Unfortunately, such systems are time-consuming and cumbersome especially when considering home environments and non-expert users. 
\begin{figure}[tp]
   \centering
      \includegraphics[scale=0.32]{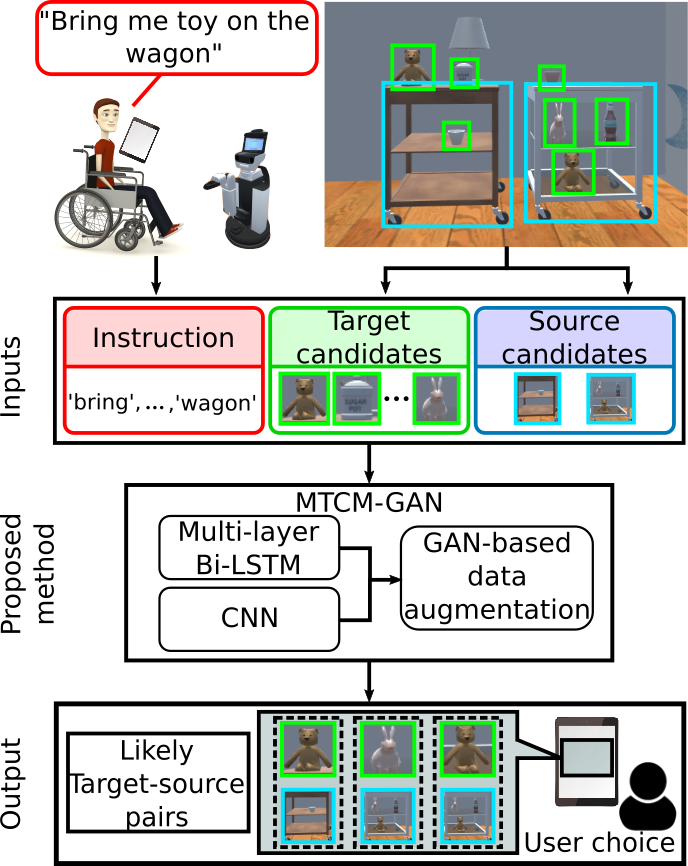}
      \caption{\small High level architecture of our MTCM-GAN that solves multimodal language understanding for fetching instructions.}
   \label{fig:architecture}
   \squeezeup
   \squeezeup
   \squeezeup
 \end{figure}
Alternatively, recent studies have combined visual and linguistic knowledge by taking a multimodal similarity-based integration approach, which uses cosine similarity between linguistic and visual information \cite{nagaraja2016modeling, yu2017joint, hatori2018interactively, Shridhar-RSS-18}.
In this approach, visual and linguistic inputs are handled by convolutional neural networks (CNNs) and long short-term memory (LSTM). Inspired by these studies, we develop a method able to understand unconstrained instructions, and predict the likelihood of candidate target objects given a single instruction. 

In this paper, we propose the multimodal target-source classifier model (MTCM) for MLU-FI tasks. MTCM is an extension of the multimodal similarity-based integration approach. Multimodal region-wise classification is used instead of cosine similarity. To handle linguistic information, we adopt sub-word embedding using BERT\cite{devlin2018bert} and a multi-layer bidirectional LSTM (Bi-LSTM). \Update In addition, we propose MTCM-GAN that extends MTCM with generative adversarial nets (GANs\cite{goodfellow2014generative}). By taking advantage of the generation ability of GANs, we show that MTCM-GAN can successfully augment data and improve on the classification performance of the  MTCM.  \Done The high level architecture of our method is shown in Fig. \ref{fig:architecture}.

The main contributions of this paper are summarized as:
\begin{itemize}
\item[$\bullet$] We propose an MTCM that predicts all target and source pairs given an instruction sentence and a scene. Rather than multimodal similarity-based integration\cite{nagaraja2016modeling, yu2017joint, hatori2018interactively, Shridhar-RSS-18}, our method is based on multimodal region-wise classification. The method is detailed in Section \ref{sec:method}. 
\item[$\bullet$] We propose MTCM-GAN, which combines the MTCM with latent classifier GAN. This is an extension of our previous work\cite{magassouba2018multimodal}. We show that the proposed method improves the classification performance in Section \ref{sec:exp}.
\end{itemize}

\section{Related work}

Inferring a user's intention relies on not only  linguistic inputs but also  other proprioceptive senses and a contextual knowledge. Several studies in the field of robotics focused on grounded communication with robots. 



Like the authors of many studies in the field of robotics, we are interested in fetching tasks in daily-life environments. Recent studies have handled multimodal language understanding using multimodal similarity-based integration\cite{nagaraja2016modeling, yu2017joint, hatori2018interactively, Shridhar-RSS-18}.
The approach proposed in \cite{nagaraja2016modeling} uses an LSTM to learn the probability of a referring expression, while a unified framework for referring expression generation and comprehension was proposed in \cite{yu2017joint}, and introduced to robotics in \cite{hatori2018interactively}.  Method handling unconstrained spoken language instructions with dialogues was proposed in \cite{hatori2018interactively}, while a robot system that comprehends human natural language instructions to pick and place everyday objects was presented in \cite{Shridhar-RSS-18},.

Unlike the above studies, we propose a region-wise classification approach, which offers more flexibility in non-fixed observations. Indeed, these studies assume a fixed observable scene for the pick-and-place task ({\it i.e.}, the target object is always in the scene), while our work has a broader DSR context. There are several cases where our region-wise classification approach has advantages over the similarity-based approach, which aims to solve the problem as a $k$-class classification problem. For example, observable scenes may not contain the target object, or erroneous instructions may be given. In such cases, the appropriate output is to classify all candidates as unlikely. Furthermore, in the case that several candidates correspond to the instruction, the region-wise classification can output multiple likely candidates to the user who may select one afterwards. For comparison purposes, both similarity-based and region-wise evaluations are presented in this paper. 

\Update We also propose a target-source network architecture in contrast with \cite{hatori2018interactively} that uses two different networks to predict the target and destination. In the latter work, the destination is predicted only to perform the placing task. In the MTCM, the source prediction is not explicitly used but is rather exploited to improve the target prediction (see Section \ref{sec:exp}). Indeed, our intuition is that such an architecture improves the target prediction by discriminating target candidates on unlikely sources. 
\Done
GANs have recently spurred the field of image reconstruction and enhancement \cite{ledig2016photo, denton2015deep}.  Interestingly, GAN-based approaches have also been used to address classification problems \cite{springenberg2015unsupervised, sugiura2018grounded, odena2016conditional, bousmalis2018using}.  The latter cited studies proposed to improve the  classification task by exploiting the data augmentation property of a GAN. In our previous works \cite{sugiura2018grounded}\cite{magassouba2018multimodal}, we introduced simultaneous data augmentation and classification to multimodal language understanding for manipulation tasks.

\section{Problem Statement}\label{sec:prob}

\begin{figure}[tp]
  \centering
   \subfloat[Bring me the empty bottle from the right wooden table]{\includegraphics[width=4. cm, height=3.5cm]{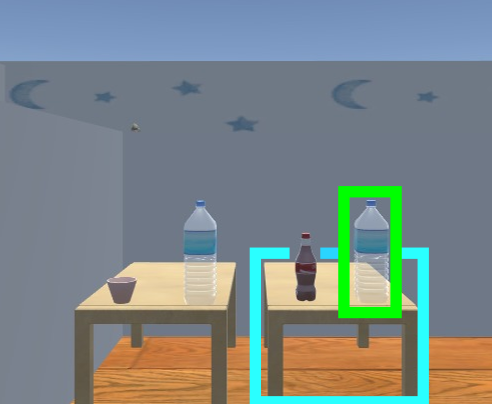}}\enskip
    \subfloat[Move the object with the black lid to the top left box]{\includegraphics[width=4. cm, height=3.5cm]{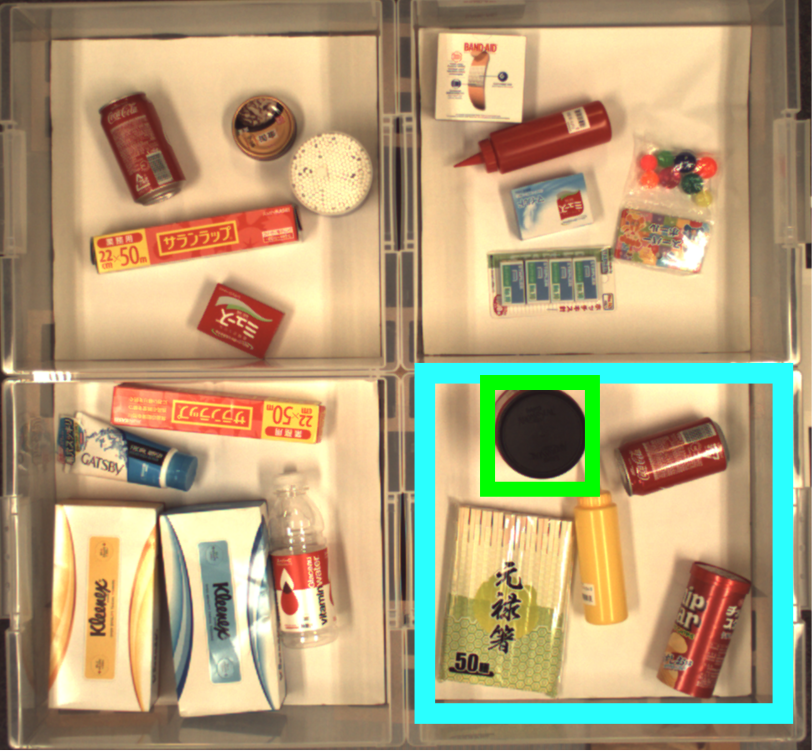}}
    \caption{\small Source (blue) and target (green) samples of the WRS-VS data set (left) and PFN-PIC data set (right). } \label{fig:sigverse} 
\squeezeup
\squeezeup
\end{figure}

\begin{figure*}[t]
   \centering
      \includegraphics[scale=0.25]{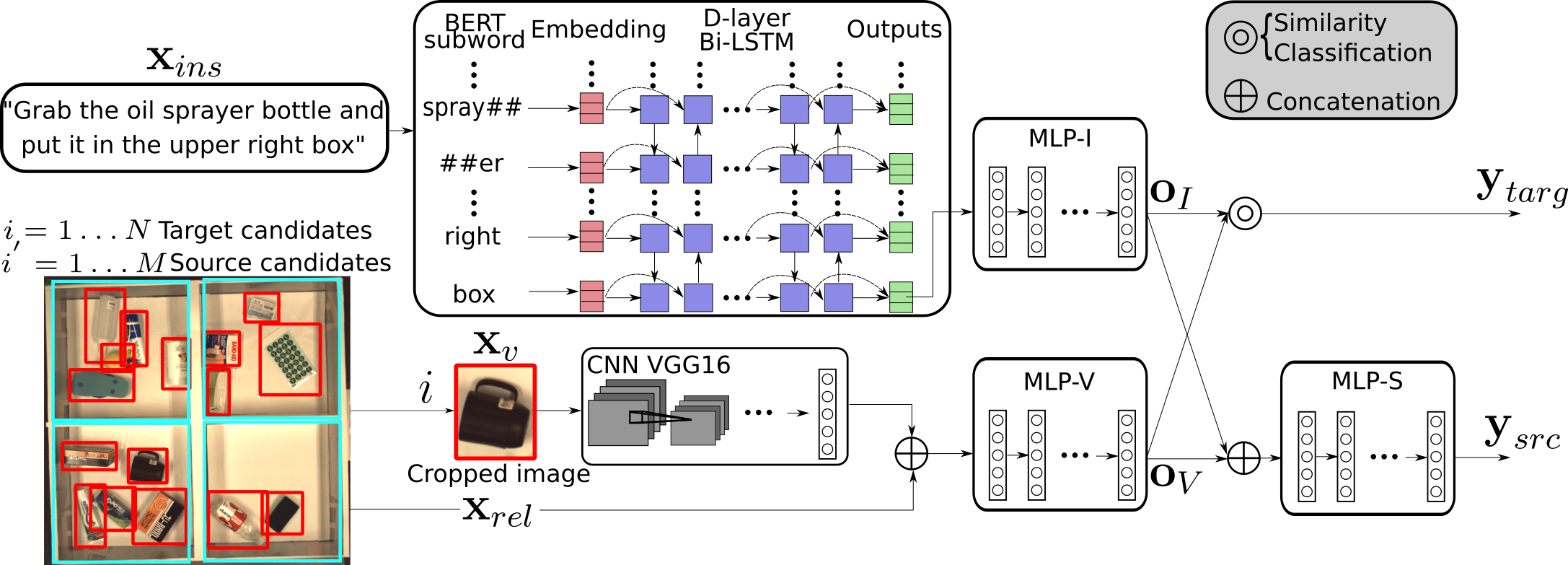}
      \caption{Proposed method framework: MTCM is composed of a multilayer Bi-LSTM, CNN, and three MLPs. The model predicts the region-wise likelihood of each target-source pair.
      Under standard conditions, the region-wise classification is used for the procedure represented by `$ \circledcirc $', while similarity-based matching is used for top-1 accuracy evaluation.}
   \label{fig:method}
  \squeezeup
\squeezeup
 \end{figure*}
\subsection{Task Description}
The present study targets MLU-FI. Figure \ref{fig:sigverse} shows typical setups of MLU-FI tasks. In an MLU-FI task, the most likely target object should be selected, given an unconstrained instruction and a scene. Typical instructions are ``Take the blue sandal and move it to the lower-left box'' and ``Bring me the soda bottle on the shelf''. Without any additional dialog, the system has to select the most likely region in the scene. This study assumes the following input and output.
\begin{itemize}
      \item[$\bullet$]{\bf Input}: An instruction sentence and a scene as an image.
      \item[$\bullet$]{\bf Output}: Likelihood of the target-source pair.  
\end{itemize}
The input and output are more thoroughly detailed in Section \ref{sec:method}. The evaluation metric is the classification accuracy. The terms {\it target} and {\it source} are defined as follows.
 \begin{itemize}
     \item[$\bullet$]{\bf Target}: The daily object (e.g. soda or fruit) that the user intend the robot to fetch 
     \item[$\bullet$]{\bf Source}: The origin of the target, such as a table or shelf
\end{itemize}
Unlike our previous work\cite{magassouba2018multimodal}, we focus on the prediction of target-source pairs, and not on {\it destination}. The definitions\footnote{Fillmore's case grammar defines the target as ``object''. We use ``target'' instead to avoid confusion because ``object'' often means a physical object in the robotics context  and the object of a sentence in the linguistics context.
} of these terms are inspired by Fillmore's case grammar\cite{fillmore1968case}.
We assume that no additional dialog-based disambiguation is used. 

MLU-FI tasks are challenging for the following reasons. First, several target and source candidates might  exist in the observable scene. The candidates may be of different or similar type. Second, users tend to use referring expressions to describe the target and/or source. Third, the expressions might be implicit, erroneous, or missing. One or more of these challenges may appear in a single instruction. We suppose that both the target and source are visible and can be detected by the robot. 

In the MLU-FI task, we assume that no additional dialog is allowed for further disambiguation. 
This is due to dialogue-based disambiguation introducing the following limitations.
\begin{itemize}
      \item[$1$] Once top-k candidates are predicted and the prediction is sufficiently accurate, a touch panel is a more reliable and efficient solution than dialog. 
      \item[$2$] Allowing dialogue-based disambiguation increases the time cost in the evaluation process because a statistically significant number of users are required to test the baseline and proposed methods. It would therefore be impractical to test many different parameters while keeping the results reproducible.
      \item[$3$] Dialogue may be unnecessary in a typical setup. For example, \cite{hatori2018interactively} reported that the validation-set accuracy was 88\% for the PFN-PIC data set, which we reproduced. This fact means that approximately 100 out of 898 validation samples are misclassified. We analyzed the errors and found that less than 20 samples among the misclassified samples required dialog-based disambiguation. This indicates that users gave clear instructions that refer to a unique target object.
\end{itemize}
We therefore focus on improving MLU accuracy.

\subsection{Task Environments}
The solution to the MLU-FI tasks should be general enough to be used for various scenarios. We therefore consider a fixed-observation configuration in the real world and a dynamic-observation configuration in simulation.

\subsubsection{Fixed observation}
We evaluate first our system on the PFN-PIC data set \cite{hatori2018interactively}.  Although PFN-PIC is designed for pick-and-place tasks, a target-source pattern can be applied to such a data set. In this data set, from a top-view, several target candidates are randomly placed into four boxes (see Fig. \ref{fig:sigverse}). These boxes are considered as source candidates. Annotated pick-and-place sentences such as ``{\it Grab the black mug and put it in the lower right box.}''  are then used. It is noted that we do not solve the placing task.

\subsubsection{Dynamic observation}
In addition to the above data set, we built an original data set using a simulator. The main advantage of this simulated environment is the wide variety of situations that can be set at a relatively small cost. In this study, we use the simulated environments that are provided in the World Robot Summit 2018 Virtual Space (WRS-VS) challenge. The simulator is based on SIGVerse \cite{inamura2013development}, which is a three-dimensional environment based on the Unity engine allowing the simulation of interactions between agents and their environments. 
In this environment, we use a simulated version of HSR (Human Support Robot), a service robot developed by Toyota having the ability to manipulate objects. 
The WRS-VS consists of typical indoor environments as illustrated in Fig. \ref{fig:sigverse}, from which we built the data set. 
Given this context, our method should understand instructions such as ``{\it Give me the green can from the table}'' or  ``{\it Take the empty bottle from the wooden table on the right side}''. 
\squeezeup


\section{Proposed method}\label{sec:method}

\Update
\subsection{Multimodal integration model}
This paragraph explains the typical approach of multimodal integration.


As illustrated in Fig. \ref{fig:method}, for each target candidate $i \in \{1, ..., N\}$ and associated source $i^{'} \in \{1,...,M\}$, we assume that their respective cropped image and positions are available. 
Thus, given a target candidate, the set of inputs ${\bf x} (i)$ is as follows:
\begin{equation}\label{equ:x}
    {\bf x}(i)=\{{\bf x}_{ins}(i), {\bf x}_v (i), {\bf x}_{rel} (i) \},
\end{equation}
where ${\bf x}_{ins}(i)$, ${\bf x}_v(i)$, and ${\bf x}_{rel}(i)$ respectively denote linguistic, visual, and relational features. Hereinafter, for readability, we voluntarily omit the index $i$, so that ${\bf x} (i)$ is written as ${\bf x}$  when further clarity is not required.

The input ${\bf x}_{rel}$ characterizes the relational features of the target candidate and environment ({\it e.g.}, other objects, location in the scene, location with respect to the source). Depending on the data set, ${\bf x}_{rel}$ may vary. Therefore, ${\bf x}_{rel}$ is explicitly given in Section \ref{sec:exp} where data sets for different tasks are introduced. 

A visual input ${\bf x}_v$ corresponds to the cropped image of the target object. A convolutional neural network (CNN) is used to process ${\bf x}_v$. 
In parallel, ${\bf x}_{ins}$ is embedded and then encoded by an LSTM network.
After encoding ${\bf x}_{ins}$ and ${\bf x}_v$, a common latent representation is required to compare the features extracted from the CNN and LSTM. Two multi-layer perceptrons (MLPs) are used for this purpose. Most existing methods use a similarity-based loss function, which is explained in detail in the appendix.
\Done



\subsection{Novelty}

We propose the MTCM. The MTCM is illustrated in Fig. \ref{fig:method} and has the following features.
\begin{itemize}
\item The MTCM is based on multimodal region-wise classification and predicts the likelihood for all candidate regions. Unlike previous methods\cite{nagaraja2016modeling, yu2017joint, hatori2018interactively, Shridhar-RSS-18}, the MTCM does not use a similarity-based loss function. 

\item We introduce a sub-word embedding model with a multi-layer Bi-LSTM to multimodal language understanding methods

\item We introduce GAN-based simultaneous data augmentation and classification to the MTCM, which is enabled by multimodal region-wise classification. Similarly to our previous method\cite{magassouba2018multimodal}, MTCM-GAN can augment data in a latent space.

\end{itemize}



\subsection{Multimodal Target-Source Classifier Model}

\Update
First, the importance of the region-wise classification model should be emphasized. A multimodal region-wise classification is more suitable to MLU-FI than the classic multimodal integration model. Indeed, in dynamic configurations (as opposed to static scenes used in \cite{yu2017joint,hatori2018interactively,Shridhar-RSS-18}), the target candidate might not be in the  scene observed, or several likely target candidates may match a given instruction. Such cases are not captured by multimodal integration models. These models are designed for static scenes and assume that there is always a unique solution in the current observed scene by the robot. This might be true in the case of image understanding \cite{yu2017joint} but not for robots in home environments. This is illustrated by  several examples in Section \ref{sec:exp}. \Done

Given the input ${\bf x}$ in Equation (\ref{equ:x}), the task is to predict the likelihood of each target-source candidate from an unconstrained instruction and a scene.  

The framework of MTCM is illustrated in Fig. \ref{fig:method}. A sub-word embedding model with a multilayer Bi-LSTM and a CNN are used for encoding linguistic and non-linguistic features.  They are connected with three MLPs, and the region-wise likelihood of the target and source is predicted.

For linguistic processing, instead of a word-based embedding model, we use a sub-word embedding model, BERT \cite{devlin2018bert}, to initialize the embedding vectors. BERT is a language encoding model based on bi-directional transformers. This approach has more flexibility and robustness. BERT was pre-trained on $3.5$ billion words and  is therefore robust against data sparseness regarding rare words. Additionally, instead of a word-based tokenization, BERT uses a sub-word tokenization\cite{wu2016google}.  The sub-word tokenization is more robust against the misspelling of words, as illustrated in Table \ref{tab:word}. 
The embedding model is then fine-tuned on the data set as the MTCM is trained.
\begin{table}[t]
\small
\caption{\small Difference between (a) typical word-tokens with pre-processing for rare and/or misspelled words and (b) sub-word tokenization.}
\label{tab:word}
\centering
\begin{tabular}{l|c|c}
\hline
Expression &(a) &(b)\\
\hline
\hline
 topright object &topright, object& top, right, object \\
\hline  
sprayer &  <UNK> & spray, er\\
\hline
greyis bottle & <UNK> , bottle & grey, is, bottle \\
\hline
\end{tabular}
\squeezeup
\squeezeup
\end{table}

We use a multi-layer Bi-LSTM instead of a simple unidirectional LSTM to encode linguistic features. In parallel, a 16-layer network VGG16 \cite{simonyan2014very} is used to encode visual features. 
These networks are connected to two MLPs; {\it i.e.}, MLP-I and MLP-V.
The output of MLP-I and MLP-V are used to predict the likelihood of a target. The outputs of the two MLPs are denoted ${\bf o}_V$ for visual features and ${\bf o}_I$ for non linguistic features.

The source is predicted by another MLP, {MLP}-{S}, based on ${\bf o}_V$ and ${\bf o}_I$.
The output of the MTCM is thus
\begin{equation}
    Y = \{ {\bf y}_{targ}, {\bf y}_{src} \},
\end{equation}
where ${\bf y}_{targ}$ and ${\bf y}_{src}$ are  respectively the target and source predictions.
MTCM's cost function $J_{MTCM}$ is defined as follows:
\begin{align}\label{equ:J_MTCM}
   	J_{MTCM} &= \lambda_1  J_{targ} + \lambda_2 J_{src}, 
\end{align}
where $\lambda_1$ and $\lambda_2$ are weighting parameters, and $J_{targ}$ and $J_{src}$ are  respectively the cross-entropy loss functions of the target and source. Given ${\bf y}$ and ${\bf y}^{*}$ as generic notations for ${\bf y}_{targ}$ and ${\bf y}_{src}$, the cross-entropy loss function $J$ is defined as 
\begin{align} \label{equ:J}
    J &= -\sum_n \sum_{m} y^{*}_{nm} \log p(y_{nm}),
\end{align}
where $y^{*}_{nm}$ denotes the label given to the $m$-th dimension of the $n$-th sample, and $y_{nm}$ denotes its prediction.


\begin{figure}[tp]
  \centering
    \includegraphics[scale=0.35]{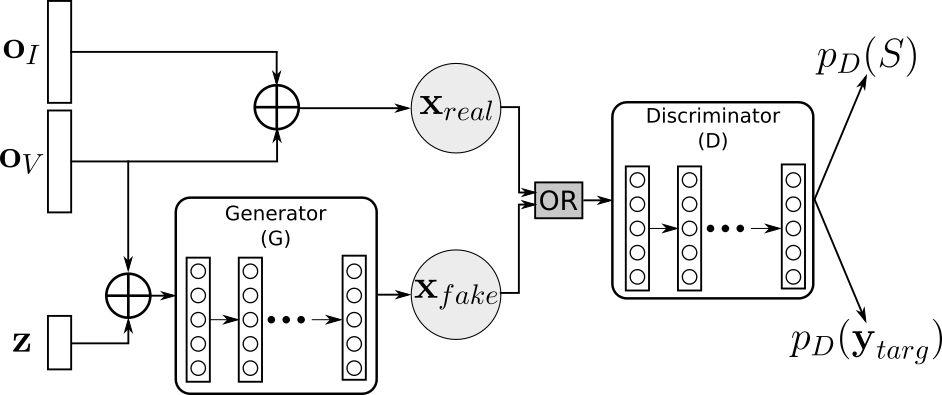}
    \caption{\small GAN-based simultaneous data augmentation and classification. The features to augment are ${\bf x}_{real}=({\bf o}_I,{\bf o}_V)$.  $G$ generates the augmented data ${\bf x}_{fake}=G({\bf z},{\bf o}_V)$. $D$ discriminates ${\bf x}_{real}$ from ${\bf x}_{fake}$ and predicts simultaneously the target likelihood $p_D({\bf y}_{targ})$. }
    \label{fig:GAN}
\squeezeup
\squeezeup
\end{figure}
\subsection{GAN-Based Simultaneous Data Augmentation and Classification}

\subsubsection{A reminder about GAN}
The GAN framework \cite{goodfellow2014generative} consists in two  adversarial networks, a discriminator $D$ and a generator $G$. The generator $G$ creates artificial data by mimicking a given data distribution. In parallel, the discriminator $D$ predicts whether the input data are real or fake. With their adversarial objectives,  $G$ is trained to generate data that are more realistic, while the discrimination ability of $D$ is enhanced.

$G$ has with a multi-dimensional input ${\bf z}$ randomly sampled from a normal distribution and generates ${\bf x}_{fake}$: 
$ {\bf x}_{fake}= G({\bf z})$.
To classify the real data ${\bf x}_{real}$ and fake data ${\bf x}_{fake}$, $D$ is alternately input with ${\bf x}={\bf x}_{real}$ or ${\bf x}={\bf x}_{fake}$ from a source flag $S \in \{real,fake\}$. The output of $D$ is $p_D(S=real|{\bf x})=D({\bf x})$.
The loss functions of $G$ and $D$ to be optimized, respectively $J_G$ and $J_D$, are defined by: 
   \begin{equation}\label{equ:J_S}
   	J_S=-\frac{1}{2} \mathbb{E}_{{\bf x}_{real}} \log D ({\bf x}_{real}) - \frac{1}{2} \mathbb{E}_{\bf z} \log(1-D({\bf x}_{fake})),
   \end{equation} 
 which results in $J_D=J_S$ and $J_G=-J_S$.

\subsubsection{From MTCM to MTCM-GAN}
In our previous works \cite{sugiura2018grounded}\cite{magassouba2018multimodal}, we introduced simultaneous data augmentation and classification that takes advantage of the data augmentation property of GANs. Indeed, artificial data generated by $G$ can be used to augment and improve a classifier network in $D$. Hence, not only does $D$ discriminate  ${\bf x}_{real}$ from ${\bf x}_{fake}$, but also the network performs a classification task by predicting the likelihood of a candidate target. Hence, in addition to $p_D(S)$, $D$ has a second output $ p_D({\bf y_{targ}})$, which is the likelihood of the target. The cost function of $D$ is then modified as $J_D=J_S+ \lambda J$
 where $\lambda$ is a weighting parameter and $J$ is the cross-entropy loss function as defined in \eqref{equ:J}.

Considering the initial MTCM network, the set of inputs of the $D$ network as illustrated in Fig.\ref{fig:GAN} is given by \begin{equation}
    {\bf x}_{GAN}=\{{\bf x}_{real}=({\bf o}_V,{\bf o}_I), {\bf x}_{fake}=G({\bf z}, {\bf o}_V)\}.
\end{equation}
\Update
Unlike the unsupervised encoding of the linguistic used in  \cite{magassouba2018multimodal}, MTCM-GAN  is a fully supervised method. Moreover, we propose an architecture  where $G$ is conditioned by ${\bf o}_V$, {\it i.e.}, ${\bf o}_V$ is input to both $D$ through the real samples and $G$. Although there is some overlap between these works MTCM-GAN can be considered an extension of the GAN method used in  \cite{magassouba2018multimodal}.
\Done
Both $D$ and $G$ are fully connected networks. Beside predicting the origin of $S$, the classifier $D$ also predicts if a pair of features $({\bf o}_I,{\bf o}_V)$ is correct, while $G$ augments the data by generating realistic pairs of correct/incorrect $({\bf o}_I,{\bf o}_V)$ features. 
To do so, similarly to the MTCM, we build pairs of correct features $({\bf o}_I(i),{\bf o}_V(i))$ and incorrect $({\bf o}_I(i),{\bf o}_V(j))$ for each target $i$  by considering a random incorrect target $j$ of the same scene. The effect of the number of incorrect pairs is studied in Section \ref{sec:exp}.

\section{Experiment (1): Static Observation}\label{sec:exp}

\subsection{Experimental Setup}
In Experiment (1), we applied the MTCM to the PFN-PIC data set that exhibits static observation conditions. 
The parameter settings are summarized in Table \ref{tab:param}.

We used the 24-layer pre-trained BERT for sub-word tokenization considering uncased words. The size of the embedded vector was 1024. We used the VGG16 pre-trained model as the CNN shown in Fig. \ref{fig:method}, and extracted the output of the seventh fully connected (FC7) layer. In MLP-I and MLP-V, we applied batch normalization and a ReLU activation function for each layer. In MLP-S, an ReLU activation function was used except for the last layer,  for which a softmax function  was used.

\begin{table}[t]
\squeezeup
\squeezeup
\small
\centering
\caption{\small Parameter settings and structures of MCTM-GAN}\label{tab:param}
\begin{tabular}{|c|l|}

\hline
MTCM Opt. & Adam (Learning rate= $2e^{-5}$, \\
method  &$\beta_1=0.99$, $\beta_2=0.9$) \\
\hline
Bi-LSTM   &$3$ layers, $1024$-cell\\
\hline
Num. & MLP-I and MLP-V: $1024$, $1024$, $1024$ \\
nodes & MLP-S: $2048$, $1024$, $128$ \\
\hline
Weight   &$\lambda_1=1, \lambda_2=0.7$ \\
\hline
\hline
GAN Opt. & Adam (Learning rate= $0.0002$, \\
method  &$\beta_1=0.5$, $\beta_2=0.9$), $\lambda=0.2$ \\

\hline
 Num. & G: $100$, $100$, $100$, $100$ \\
 nodes.   & D:  $100$, $200$, $400$, $1000$ \\
\hline
\hline
 Batch sizes&$64$ (for $G$ and $D$)  and $128$ (for MTCM)  \\
\hline
\end{tabular}
\squeezeup
\squeezeup
\end{table}

\begin{figure*}[t]
   \captionsetup[subfigure]{justification=raggedright, width = 1.8cm }
   \parbox{\LW}{\includegraphics[width=\linewidth-1mm]{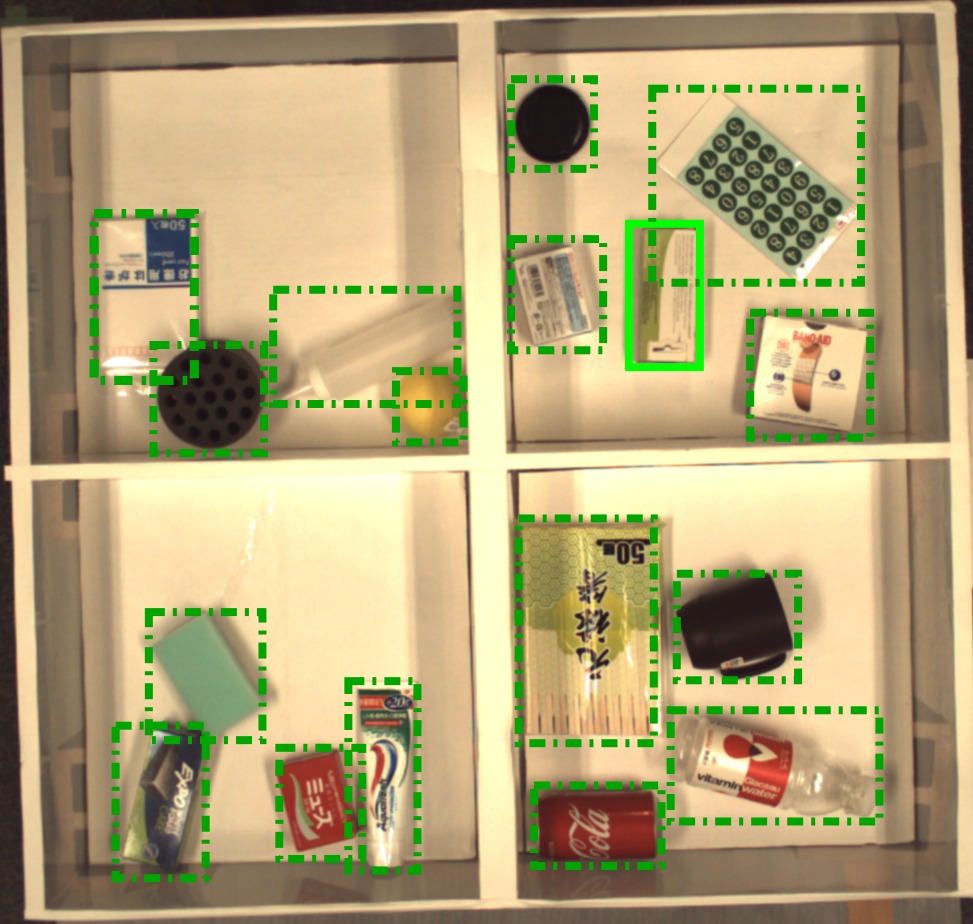}}
   \parbox{\LWb}{\subfloat[Take the white and green object in center of upper right box and place it in the upper left box.]}\qquad\qquad
  \parbox{\LW}{\includegraphics[width=\linewidth-1mm]{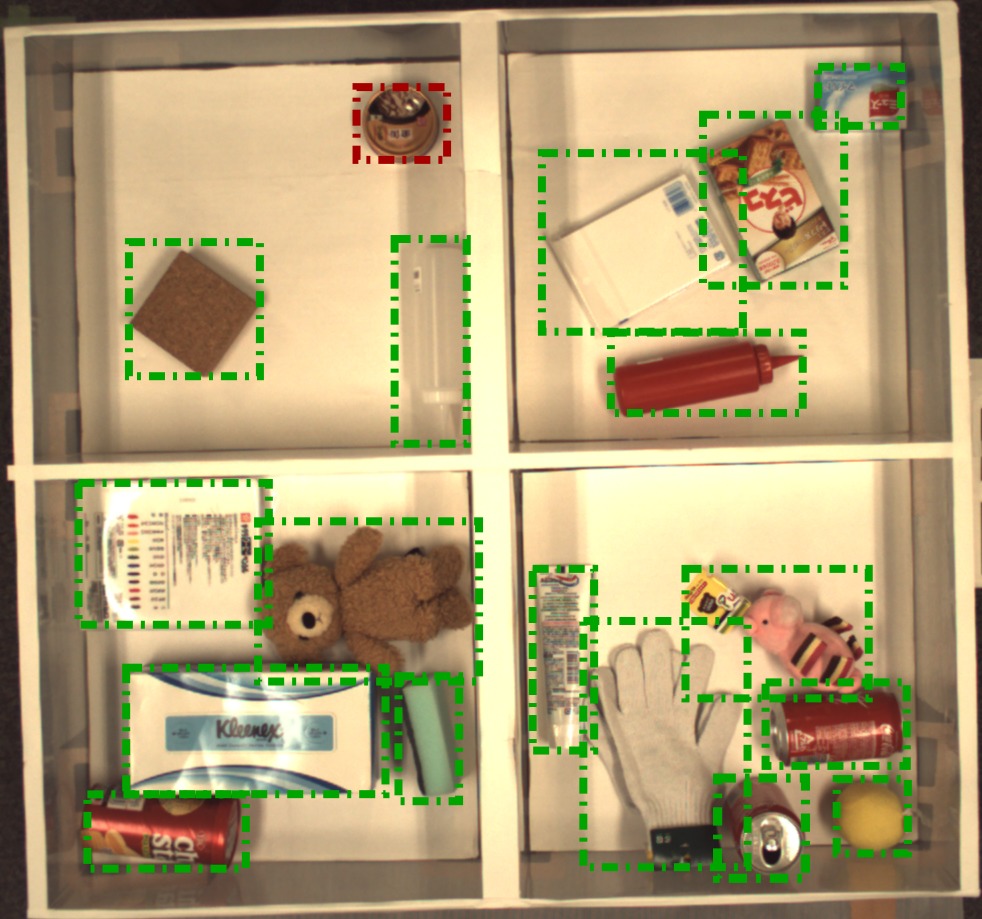}}
  \parbox{\LWb}{\subfloat[Move the large brown coin or button to bottom left box.]}\qquad\qquad
  \parbox{\LW}{\includegraphics[width=\linewidth-1mm]{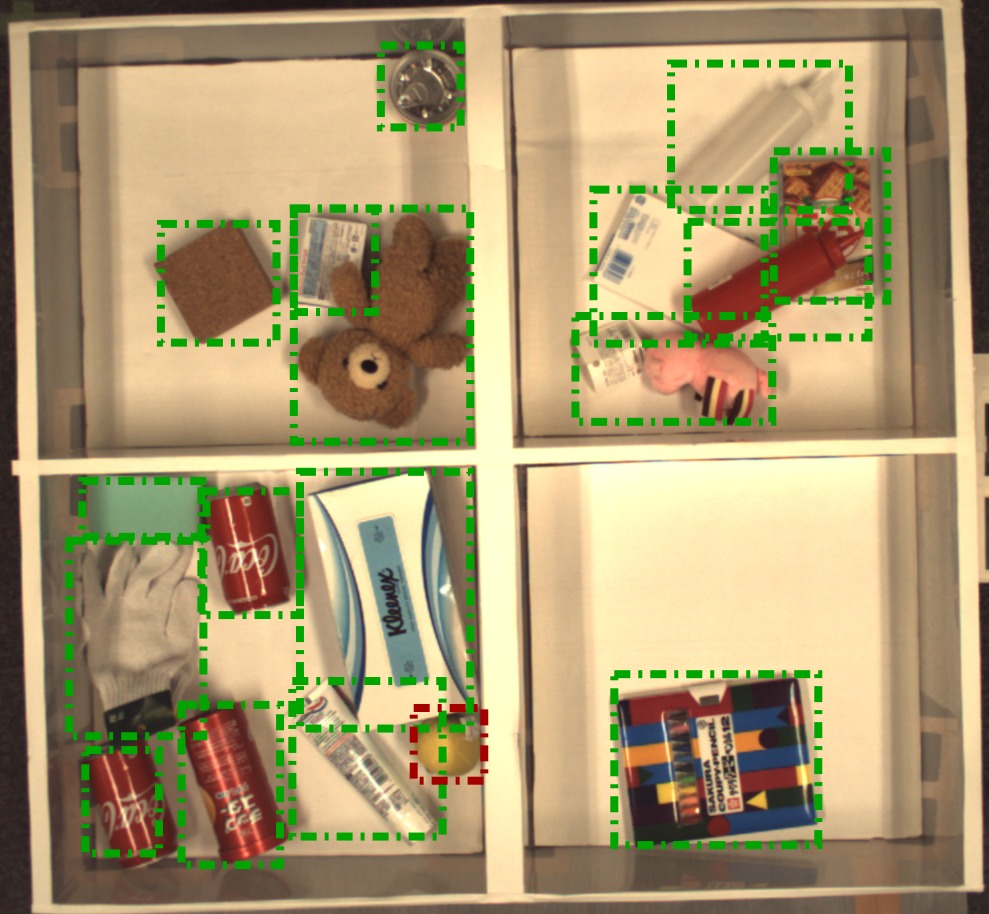}}
  \parbox{\LWb}{\subfloat[Take the green ball from lower left box to lower right box.]}\\
   \parbox{\LW}{\includegraphics[width=\linewidth-1mm]{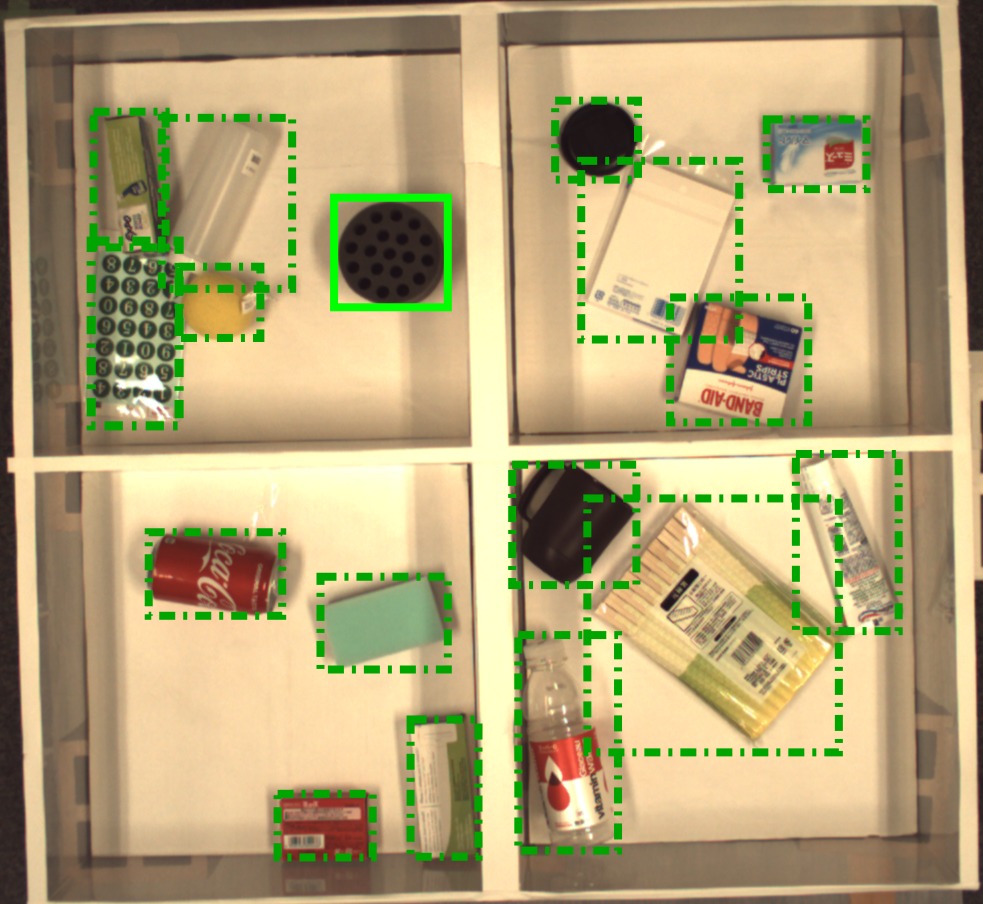}}
   \parbox{\LWb}{\subfloat[Move the larger black circular object into the lower left box.]}\qquad\qquad
  \parbox{\LW}{\includegraphics[width=\linewidth-1mm]{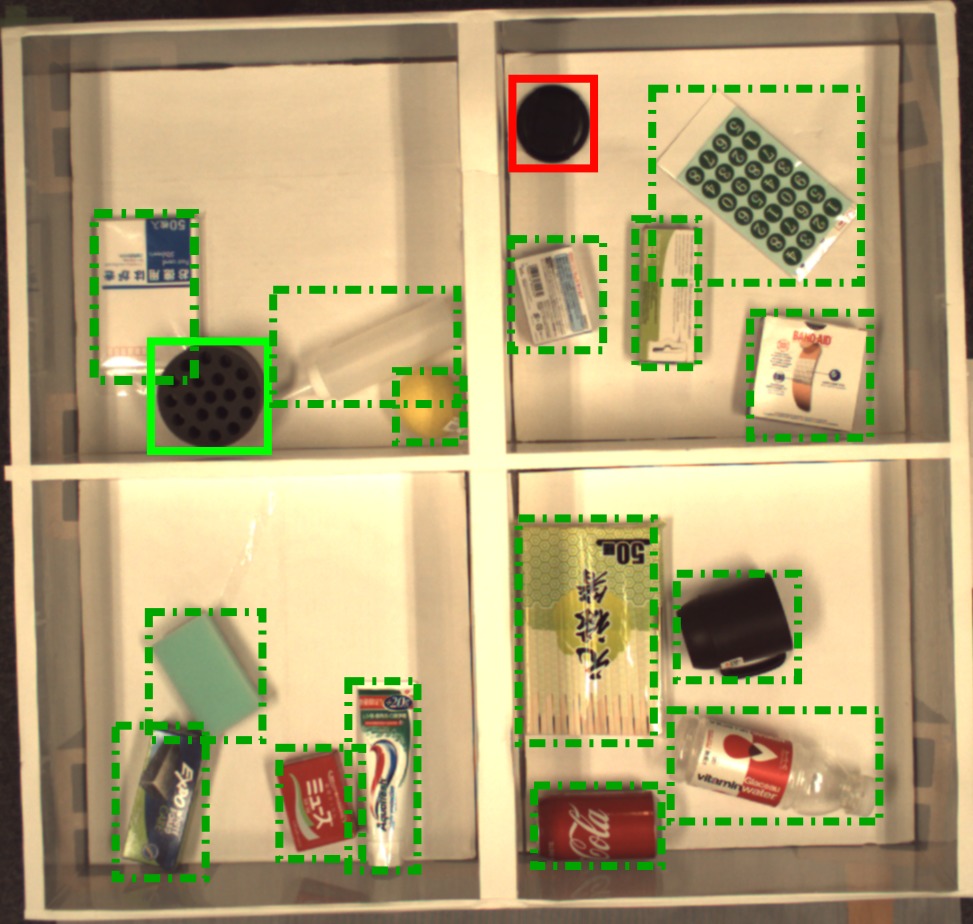}}
  \parbox{\LWb}{\subfloat[Grab the black round thing from the top left box and put in the lower right box.]}\qquad\qquad
  \parbox{\LW}{\includegraphics[width=\linewidth-1mm]{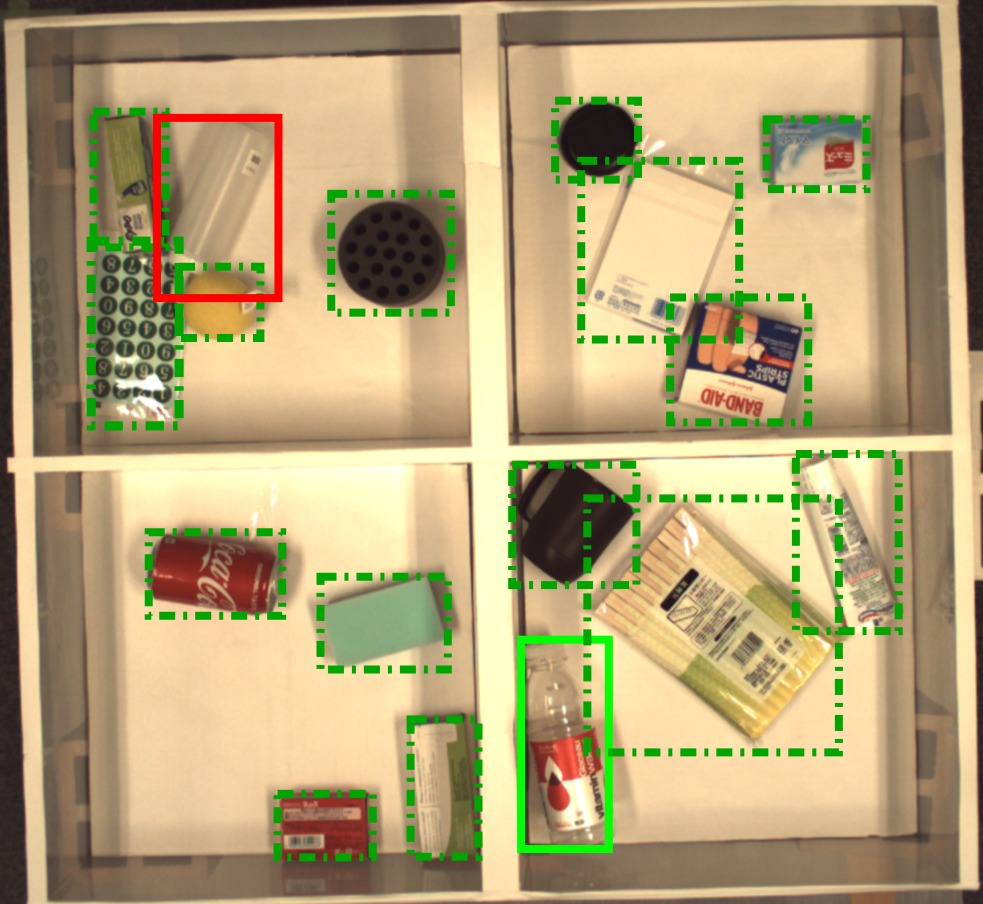}}
  \parbox{\LWb}{\subfloat[Pick the white plastic bottle and put it in the right box.]}

    \caption{\small \Update Predictions made using our method. Solid and dotted rectangles colored  green and dark green, respectively, represent true positives and negatives. In contrast, solid and dotted rectangles colored red and dark red, respectively, represent false positives and negatives. The left subfigures (a) and (d) show correct predictions, the middle subfigures (b) and (e) show misclassified samples, and the right subfigures (c) and (f) show misclassified samples due to the ambiguous or erroneous instructions. \Done}
    \label{fig:samples}
   \squeezeup
  \squeezeup
  \squeezeup
\end{figure*}

The generator $G$ and discriminator $D$ both comprised of four layers with ReLU activation functions. Batch normalization was applied to these layers. The output layer of $G$ was a tanh activation function, while a softmax function was applied to the output layer of $D$. As described previously in Section \ref{sec:method}, $G$ is conditioned by ${\bf o}_V$  so that the input of $G$ is defined by (${\bf z},{\bf o}_V$),  where ${\bf z}\sim \mathcal{N}(0,1)$ with a dimension $d_z=100$. The input dimension of $G$ is $d_G= 1124$. The output of $G$ has a dimension $d_{fake}=2048$.

\subsection{Data Set: PFN-PIC Data Set}

Although there are multiple versions of the PFN-PIC data set, we used the same version reported in \cite{hatori2018interactively} : $89,891$ sentences in the training set and $898$ sentences in the validation set.
The PFN-PIC data set was annotated by three annotators, and contains 1,646 unique words. The average length of sentences is 14 words.

To solve the MLU-FI task with the PFN-PIC data set,  we define the relational feature ${\bf x}_{rel}(i)$, for a candidate target $i$ as its position and size in the image. Similarly to \cite{yu2017joint}, we express the relational feature as ${\bf x}_{rel}=\begin{bmatrix}
\frac{x}{W}, & \frac{y}{H}, &\frac{w}{W},&\frac{h}{H},&\frac{wh}{WH}
\end{bmatrix}$, where $(x,y,w,h)$ denotes the (horizontal position, vertical position, width, height) of the target, while $W$ and $H$ are to the width and height of the scene image.

\subsection{Qualitative Results (1)}

The qualitative results shown in Fig. \ref{fig:samples} illustrate typical true and false predictions. Each column  presents the results of MTCM-GAN for different situations. Subfigures (a) and (d) on the left present correct predictions. 

Misclassification are illustrated in subfigures (b) and (e). In subfigure (b), our method misclassified the brown object in the top-left box, which was referred to as a ``large brown coin or button.'' In subfigure (e), the source is considered to be the origin of the error because the false positive example shown with the red rectangle matches the description (``black round thing''). Indeed, the false positive was  due to the statement of an incorrect box (top right instead of top left).

Subfigures (c) and (f) present candidates misclassified owing to ambiguous or erroneous instructions.
In the case of subfigure (c), the target is described as the ``green ball from lower left box'', and the ground-truth target is shown by the dotted red rectangle.
MTCM-GAN predicted all objects as unlikely, leading to a false negative for the ground-truth target. However, the ground-truth appears ``yellow'' and not ``green'' to most people. It is reasonable to regard the false-negative prediction as incorrect.
Meanwhile, for subfigure (f), the instruction was ``Pick the white plastic bottle and put it in the right box'' and the object shown by the solid green rectangle was annotated as the target that fits the instruction. However, the object shown by the solid red rectangle also looks like a ``white plastic bottle,'' and was predicted as a likely target by our method. There were thus multiple target candidates that fit the given instruction. It is  also reasonable to regard the false-positive prediction as incorrect.

The latter results emphasize an important point about the region-wise classification method, {\it i.e.}, such situations that might occur in real-world contexts are not naturally handled by similarity-based approaches, which always return the best-matching target. By contrast, our approach offers more flexibility; {\it i.e.},  it understands when the specified target is not in the scene or when several likely targets are in the scene and provides appropriate feedback to the user. This flexibility is crucial for interacting with non-expert users and an asset for DSRs.

\squeezeup

\subsection{Quantitative Results (1)}

\begin{table*}[t]
\normalsize
\caption{\small Validation of top-1 and region-wise accuracy on the PFN-PIC data set. The region-wise accuracy is also reported for several positive/negative samples ratio $\gamma$. The mean accuracy and sample standard deviation over five trials are given.}
\label{tab:results}
\centering
\begin{tabular}{l|c|ccccc}
\hline
&\multicolumn{6}{c}{\bf Target accuracy $[\%]$} \\
\cline{2-7}
{\bf Method }&\multicolumn{1}{c|}{\bf Top-1 } & \multicolumn{5}{c}{\bf Region-wise} \\
\cline{3-7}
& &\multicolumn{1}{c|}{$\gamma=1$} &\multicolumn{1}{c|}{$\gamma=2$} & \multicolumn{1}{c|}{$\gamma=4$} & \multicolumn{1}{c|}{$\gamma=5$}& \multicolumn{1}{c}{$\gamma=10$}\\
\hline
\hline
Baseline \cite{hatori2018interactively} & 88.0 &$-$&$-$ &$-$ &$-$ &$-$\\
\hline
MTCM & 88.8$\pm$0.43 & 94.5$\pm$0.22 & 95.4$\pm$0.27 & 96.1$\pm$0.17 & 96.3$\pm$0.12 & 97.1$\pm$0.06 \\
\hline
MTCM-GAN w/o conditioning &$-$ & 95.7$\pm$0.18 & 96.1$\pm$0.21 & 96.2$\pm$0.13 & 96.4$\pm$0.09 & 96.8$\pm$0.19 \\
\hline
MTCM-GAN w/ conditioning &$-$ & 95.9$\pm$0.18 & 96.3$\pm$0.12 & 96.5$\pm$0.11 & 96.5$\pm$0.07 & 97.2$\pm$0.12 \\
\hline
\end{tabular}
\squeezeup
\squeezeup
\end{table*}

Table \ref{tab:results} presents quantitative results. The top-1 accuracy of the baseline was reported in \cite{hatori2018interactively}. Other presented in the table are  the mean accuracy and sample standard deviation, over five trials.

In the first column, for a fair comparison with the baseline, we provide the top-1 accuracy defined in \cite{hatori2018interactively}. Likewise, we used the similarity-based matching (see Appendix) for the procedure shown as `$ \circledcirc $' in Fig. \ref{fig:method}. This means that we did not use the region-based classification for the top-1 accuracy. Nonetheless, the MTCM achieveda top-1 accuracy of 88.8\% while the baseline method achieved 88.0\%. Although the top-1 accuracy is not the main focus of this paper, the result indicates that the MTCM can predict the target more accurately than the baseline. \Update Note that the region-wise accuracy was not reported in \cite{hatori2018interactively}, and statistical comparison with the latter work is limited because both the average and standard deviation are needed for such an investigation. Nonetheless we report the results of an additional ablation study in Table \ref{tab:ablation} to emphasize the different contributions of the network.  The similarity-based  state-of-the-art method  used in \cite{yu2017joint} is referred as Classic.  The  table presents the effect of BERT and the source prediction. BERT increases the accuracy by 1.9\%. The sub-word model that is more robust to spelling errors and the attention mechanism used to build the models better capture  linguistic characteristics. The positive effect of BERT and more particularly the attention mechanism on the linguistic processing suggest that such a method would also benefit to the visual encoding. 
Additionally, the source prediction improves the target prediction accuracy by 3.4\%. Although the predicted source is not explicitly used, this result emphasizes the idea that a model trained to predict the likelihood of a source candidate is more likely to predict the likelihood of a target candidate. This is particularly the case for MLU-FI, where the target source may be mentioned. 
\begin{table}[t]
\normalsize
\caption{\small \Update Ablation study of the MTCM \Done}
\label{tab:ablation}
\centering
\begin{tabular}{l|c}
\hline
{\bf Method }& \multicolumn{1}{c}{\bf Top-1  $[\%]$} \\
\hline
\hline
Classic & 84.6$\pm$0.63 \\
\hline
Classic + BERT & 86.5$\pm$0.84  \\
\hline
Classic + Source & 88.0$\pm$0.73 \\
\hline
MTCM(Classic + BERT + Source) &88.8$\pm$0.43  \\
\hline
\end{tabular}
\squeezeup
\squeezeup
\end{table}
\Done
We  next compare the MTCM and MTCM-GAN sing region-wise accuracy, which is the main focus of the paper. 
From the prediction ${\bf y}$ and the ground truth label ${\bf y}^{*}$, the region-wise accuracy  $E_r$ is defined as:
\begin{equation} \label{equ:biacc}
 E_r = \sum \frac{N_{{\bf y} = {\bf y}^{*}}}{  |\{({\bf o}_V,{\bf o}_I)\}| }.
\end{equation}
For each target $i$ a set $R(i)=R_{+}(i) \cup \gamma R_{-}(i)$  of correct and incorrect feature pairs was built. More explicitly, $R_{+}(i)=\{({\bf o}_V(i),{\bf o}_I)(j)\}$  while $R_{-}(i)=\{({\bf o}_V(j),{\bf o}_I)(i)\}$, where $j$ was sampled from a random target of the same image as $i$ and $j \neq i$. The parameters $\gamma$ characterized the number of incorrect samples for each correct sample.

Table \ref{tab:results} shows that the MTCM achieved region-wise accuracies 94.5\%, 95.4\%, 96.1\%, 96.3\% and 97.1\% when $\gamma$ was $1,2,4,5$ and $10$, respectively. Note that the top-1 accuracy and region-wise accuracy cannot be compared directly. In the table, ``MTCM-GAN with conditioning'' represents the model in which $G$ is conditioned by (${\bf z},{\bf o}_V$), as described previously in Section \ref{sec:method}. MTCM-GAN outperformed then MTCM when $\gamma \in \{1,2,4,5,10\}$. 

\begin{figure}[tp]
   \centering
      \includegraphics[scale=0.4]{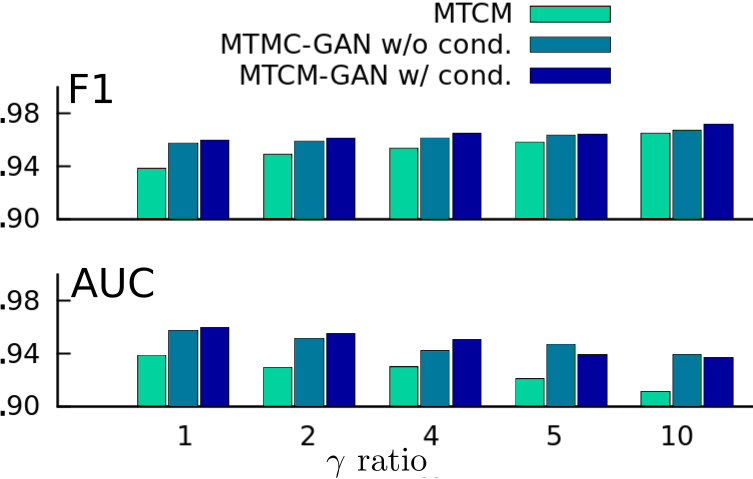}
      \caption{\small \Update F1 score (top) and AUC metric (bottom) for the MTCM and MTCM-GAN considering different ratios $\gamma$ \Done}
   \label{fig:F1}
   \squeezeup
   \squeezeup
   \squeezeup
 \end{figure}

Additionally, we investigated the effect of ${\bf o}_V$. We tested a model in which $G$ is conditioned by ${\bf z}$, referred to as ``MTCM-GAN without conditioning'' in the table. MTCM-GAN with conditioning had higher region-wise accuracy than that without conditioning under all conditions. This indicates that a more accurate prediction can be achieved by introducing ${\bf o}_V$. 
The table shows that the superiority of MTCM-GAN over MTCM was greater when $\gamma$ was smaller. For example, the improvement of MTCM-GAN over MTCM was 1.4\% when $\gamma = 1$ and 0.9\% when $\gamma = 2$. This indicates that the data augmentation property is more effective with fewer samples. Such a property is particularly interesting in robotics, in which real data collection and annotation are generally long and tedious tasks  and few large-scale data sets are available in contrast with the case in the fields of computer vision and natural language. The lack of data is compensated by this type of augmentation. 
\Update 
Furthermore, although there is no guaranteed correlation between the generated samples quality and the accuracy improvement,  MTCM-GAN can also be evaluated with generation metrics. Considering the Frechet inception distance \cite{shmelkov2018good} as the metric, MTCM-GAN generator produced samples with a score of 11.64 at the end of the training.
\Done
\Update
As illustrated in Fig. \ref{fig:F1}, the conditioning effect, as well as the superiority of MTCM-GAN over MTCM is confirmed when considering the F1 score.  The  area under the curve (AUC) metric gives similar results although the positive effect of conditioning is not guaranteed for high $\gamma$. 
\Done

\section{Experiment (2): Dynamic Observation}\label{sec:exp2}
In Experiment (2), we validated our method with the WRS-VS data set explained in Section \ref{sec:prob}. We used the same parameter settings and structures given in Table \ref{tab:param}. Except for the input dimensions and ${\bf x}_{rel}$, the characteristics of the networks are shared between Experiment (1) and (2).
\subsection{Data Set: WRS-VS Data Set}
For the WRS-VS data set, we collected $308$ images from which we could annotate $1010$ targets in the training set and $37$ targets in the validation set. The data set was annotated by an expert user. This data set has an average of $3.4$ targets per image, and $7.4$ words for each instruction. In addition to ${\bf x}_{rel}$ defined previously, we used the relative positions of the source and the target and the cropped image of the source.

\subsection{Qualitative Results (2)}

\begin{figure*}[t]
  \captionsetup[subfigure]{justification=raggedright, width = 1.5 cm }
  \parbox{\LW}{\includegraphics[width=\linewidth]{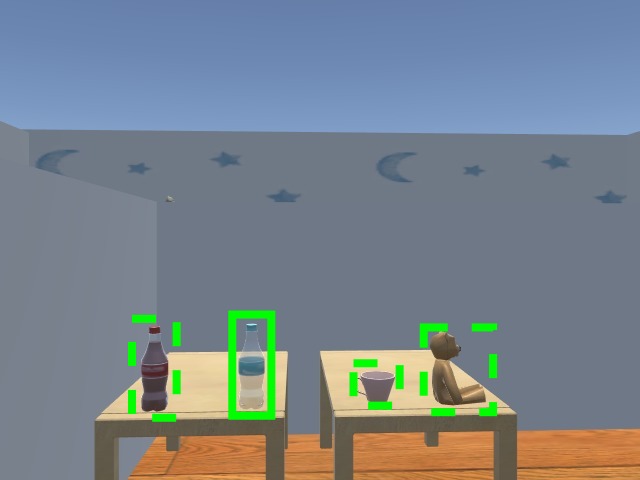}}
  \parbox{\LWb}{\subfloat[Give me the empty bottle on the left table]{}}\qquad\enskip
  \parbox{\LW}{\includegraphics[width=\linewidth]{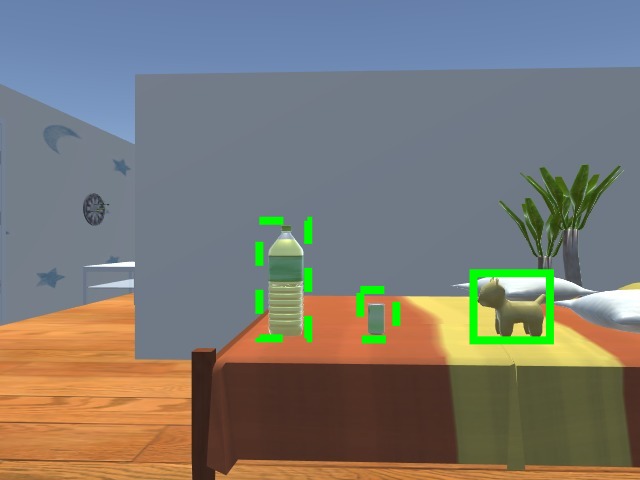}}
  \parbox{\LWb}{\subfloat[Bring me the yellow doll from the bed with orange sheets]{}}\qquad\enskip\enskip
  \parbox{\LW}{\includegraphics[width=\linewidth]{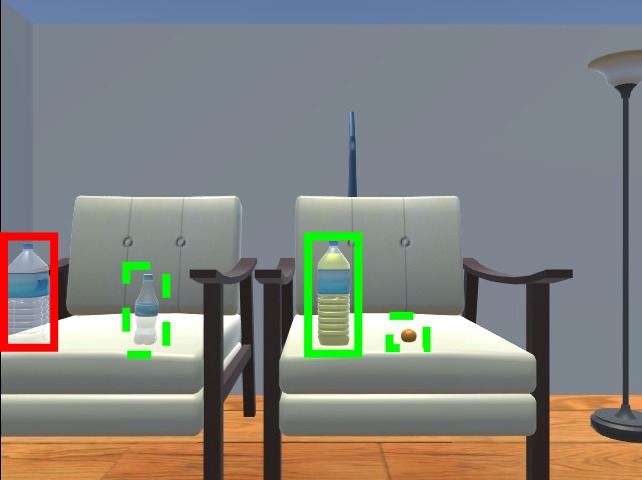}}
  \parbox{\LWb}{\subfloat[On the left sofa chair take the big bottle]}

    \caption{\small Predictions made using method. Solid and dotted rectangles colored green respectively represent true positives and negatives. In contrast, solid and dotted rectangles colored red respectively represent false positives and negatives.}
    \squeezeup
    \squeezeup
    \squeezeup
    \label{fig:wrs_samples}
\end{figure*}

Qualitative results of our method are illustrated in Fig. \ref{fig:wrs_samples}.
Subfigures (a) and (b) present correct predictions, which confirm the consistency of our method independently of the data set used. Nonetheless,  ther are erroneous predictions, as in subfigure (c) when the scene configuration is challenging.

\subsection{Quantitative Results (2)}

We also report in Table \ref{tab:results_wrs} accuracy results for the MTCM and MTCM-GAN without and with conditioning. Note that these results are obtained only for $\gamma \in \{1,2\}$ because the average number of targets in each scene ($3.4$) is low. 

Similarly to the PFN-PIC case, the results confirm the superiority of MTCM-GAN over the MTCM. Indeed, for both $\gamma =1$ and $\gamma=2$, a higher accuracy was achieved with MTCM-GAN (90.7\% and 93.6\%) than with MTCM (83.5\% and 85.6\%). 
Likewise, the conditioning effect on MTCM-GAN improves the simultaneous data augmentation and classification because  the most accurate results were obtained with this method.

\section{Conclusion}

Motivated by the increasing demand for domestic service robots, we developed a method of understanding unconstrained fetching instructions. The proposed method (MTCM) is a region-wise classifier that predicts the likelihood of target objects and their respective sources given linguistic and visual inputs. The following important contributions of the paper are emphasized:

\begin{itemize}
 \item The MTCM is based on multimodal region-wise classification, which predicts the likelihood of all candidate regions. We introduced a sub-word embedding model with  Bi-LSTM to multimodal language understanding methods. The MTCM achieved accuracy of 88.8\%, which was 0.8\% higher than the top-1 accuracy in \cite{hatori2018interactively}.
\item We introduced GAN-based simultaneous data augmentation and classification to the MTCM, which was enabled by multimodal region-wise classification. The region-wise accuracy of MTCM-GAN was better than the initial accuracy of MTCM. 
\end{itemize}

In future work, we plan to extend MTCM-GAN with attention mechanisms that have proven to work well. A physical experimental study with non-expert users and a human service robot is also planned.

\begin{table}[t]
\normalsize
\caption{\small Average region-wise accuracy on the WRS-VS data set w.r.t $\gamma$.}
\label{tab:results_wrs}
\centering
\begin{tabular}{l|cc}
\hline
{\bf Method }& \multicolumn{2}{c}{\bf Region-wise $[\%]$} \\
\cline{2-3}
 &\multicolumn{1}{c|}{$\gamma=1$} &\multicolumn{1}{c}{$\gamma=2$}\\
\hline
\hline
MTCM & 83.5$\pm$1.4 & 85.6$\pm$1.2 \\
\hline
MTCM-GAN w/o cond. & 89.6$\pm$1.6 & 92.6$\pm$1.3  \\
\hline
MTCM-GAN w/ cond. & 90.7$\pm$0.9 &  93.6$\pm$1.1  \\
\hline
\end{tabular}
\squeezeup
\squeezeup
\end{table}
\vspace{-1mm}
\appendix 

\section{Multimodal Similarity-Based Integration} 
\vspace{-1mm}
Hereinafter we explain the similarity-based loss function used in previous work\cite{nagaraja2016modeling, yu2017joint, hatori2018interactively, Shridhar-RSS-18}. 
In those similarity-based approaches, a hinge loss function ${J}_{sim}$ is used to match linguistic and non-linguistic features. This loss function consists in increasing the similarity between correct pairs of linguistic and non-linguistic features and the dissimilarity between incorrect pairs. More explicitly the loss function is expressed by
$J_{sim}=\sum_i \bigl\{
   	\max \bigl(0, \lambda_M + f(g_1(i),g_2(j) )
   	- f(g_1(i),g_2(i) \bigr)
   	+ \max \bigl( 0, \lambda_M + f(g_1(k),g_2(i) )
   	- f(g_1(i),g_2(i) \bigr) \bigr\},
$
where $\lambda_M$ is the margin, and $f(\cdot,\cdot)$ is a similarity function such as cosine similarity. The functions $g_1(\cdot)$ and $g_2(\cdot)$ correspond to the neural networks related to the linguistic and non-linguistic features, respectively. The incorrect linguistic and non-linguistic features are extracted from two random candidates targets $j$ and $k$ from the same image as the current target $i$. This means that $j$ and $k$ are randomly sampled from $\{1,...,N\}$, where $j \not= i$ and $k \not= i$.



\vspace{-2mm}
\section*{Acknowledgements}
\vspace{-1mm}
This work was partially supported by JST CREST and SCOPE.
\vspace{-3mm}

\bibliographystyle{IEEEtran}
\bibliography{bibthese}

\begin{thebibliography}{10}
\providecommand{\url}[1]{#1}
\csname url@rmstyle\endcsname
\providecommand{\newblock}{\relax}
\providecommand{\bibinfo}[2]{#2}
\providecommand\BIBentrySTDinterwordspacing{\spaceskip=0pt\relax}
\providecommand\BIBentryALTinterwordstretchfactor{4}
\providecommand\BIBentryALTinterwordspacing{\spaceskip=\fontdimen2\font plus
\BIBentryALTinterwordstretchfactor\fontdimen3\font minus
  \fontdimen4\font\relax}
\providecommand\BIBforeignlanguage[2]{{%
\expandafter\ifx\csname l@#1\endcsname\relax
\typeout{** WARNING: IEEEtran.bst: No hyphenation pattern has been}%
\typeout{** loaded for the language `#1'. Using the pattern for}%
\typeout{** the default language instead.}%
\else
\language=\csname l@#1\endcsname
\fi
#2}}

\bibitem{brose2010role}
S.~W. Brose, D.~J. Weber, \emph{et~al.}, ``{The role of assistive robotics in
  the lives of persons with disability},'' \emph{American Journal of Physical
  Medicine \& Rehabilitation}, vol.~89, no.~6, pp. 509--521, 2010.

\bibitem{iocchi2015robocup}
L.~Iocchi, D.~Holz, J.~Ruiz-del Solar, K.~Sugiura, and T.~Van Der~Zant,
  ``{RoboCup@ Home: Analysis and Results of Evolving Competitions for Domestic
  and Service Robots},'' \emph{Artificial Intelligence}, vol. 229, pp.
  258--281, 2015.

\bibitem{gemignani2015language}
G.~Gemignani, M.~Veloso, and D.~Nardi, ``{Language-based Sensing Descriptors
  for Robot Object Grounding},'' in \emph{Robot Soccer World Cup}, 2015, pp.
  3--15.

\bibitem{nagaraja2016modeling}
V.~K. Nagaraja, V.~I. Morariu, and L.~S. Davis, ``{Modeling Context between
  Objects for Referring Expression Understanding},'' in \emph{ECCV}.\hskip 1em
  plus 0.5em minus 0.4em\relax Springer, 2016, pp. 792--807.

\bibitem{yu2017joint}
L.~Yu, H.~Tan, M.~Bansal, and T.~L. Berg, ``{A joint Speaker
  Listener-Reinforcer Model for Referring Expressions},'' in \emph{CVPR},
  vol.~2, 2017.

\bibitem{hatori2018interactively}
J.~Hatori \emph{et~al.}, ``{Interactively Picking Real-World Objects with
  Unconstrained Spoken Lnguage Instructions},'' in \emph{IEEE ICRA}, 2018, pp.
  3774--3781.

\bibitem{Shridhar-RSS-18}
M.~Shridhar and D.~Hsu, ``Interactive visual grounding of referring expressions
  for human-robot interaction,'' in \emph{RSS}, 2018.

\bibitem{devlin2018bert}
J.~Devlin, M.~W. Chang, K.~Lee, and K.~Toutanova, ``{BERT}: Pre-training of
  deep bidirectional transformers for language understanding,'' \emph{arXiv
  preprint arXiv:1810.04805}, 2018.

\bibitem{goodfellow2014generative}
I.~Goodfellow, J.~Pouget-Abadie, M.~Mirza, B.~Xu, D.~Warde-Farley, S.~Ozair,
  A.~Courville, and Y.~Bengio, ``{Generative Adversarial Nets},'' in
  \emph{NIPS}, 2014, pp. 2672--2680.

\bibitem{magassouba2018multimodal}
A.~Magassouba, K.~Sugiura, and H.~Kawai, ``{A Multimodal Classifier Generative
  Adversarial Network for Carry and Place Tasks From Ambiguous Language
  Instructions},'' \emph{IEEE RA-L}, vol.~3, no.~4, pp. 3113--3120, Oct 2018.

\bibitem{ledig2016photo}
C.~Ledig, L.~Theis, F.~Husz{\'a}r, \emph{et~al.}, ``{Photo-Realistic Single
  Image Super-Resolution Using a Generative Adversarial Network},''
  \emph{arXiv:1609.04802}, 2016.

\bibitem{denton2015deep}
E.~L. Denton, S.~Chintala, R.~Fergus, \emph{et~al.}, ``{Deep Generative Image
  Models Using a Laplacian Pyramid of Adversarial Networks},'' in \emph{NIPS},
  2015, pp. 1486--1494.

\bibitem{springenberg2015unsupervised}
J.~T. Springenberg, ``{Unsupervised and Semi-supervised Learning with
  Categorical Generative Adversarial Networks},'' \emph{arXiv preprint
  arXiv:1511.06390}, 2015.

\bibitem{sugiura2018grounded}
K.~Sugiura and H.~Kawai, ``{Grounded Language Understanding for Manipulation
  Instructions Using GAN-Based Classification},'' \emph{IEEE ASRU}, 2017.

\bibitem{odena2016conditional}
A.~Odena, C.~Olah, and J.~Shlens, ``{Conditional Image Synthesis with Auxiliary
  Classifier {GAN}s},'' in \emph{ICML}, 2017, pp. 2642--2651.

\bibitem{bousmalis2018using}
K.~Bousmalis \emph{et~al.}, ``Using simulation and domain adaptation to improve
  efficiency of deep robotic grasping,'' in \emph{Proc. IEEE ICRA}, 2018, pp.
  4243--4250.

\bibitem{fillmore1968case}
C.~J. Fillmore, ``The case for case. universals in linguistic theory,''
  \emph{New York. Holt, Rinehart \& Winston. Syntax-Semantics Interface in
  Psych-verb Constructions (2341)}, vol. 201, pp. 100\,363--76, 1968.

\bibitem{inamura2013development}
T.~Inamura, J.~T.~C. Tan, K.~Sugiura, T.~Nagai, and H.~Okada, ``Development of
  robocup@ home simulation towards long-term large scale hri,'' in \emph{Robot
  Soccer World Cup}.\hskip 1em plus 0.5em minus 0.4em\relax Springer, 2013, pp.
  672--680.

\bibitem{wu2016google}
Y.~Wu \emph{et~al.}, ``Google's neural machine translation system: Bridging the
  gap between human and machine translation,'' \emph{arXiv preprint
  arXiv:1609.08144}, 2016.

\bibitem{simonyan2014very}
K.~Simonyan and A.~Zisserman, ``Very deep convolutional networks for
  large-scale image recognition,'' \emph{arXiv preprint arXiv:1409.1556}, 2014.

\bibitem{shmelkov2018good}
K.~Shmelkov, C.~Schmid, and K.~Alahari, ``{How good is my GAN?}'' in
  \emph{ECCV}, 2018, pp. 213--229.

\end{thebibliography}

\end{document}